\def\year{2021}\relax
\documentclass[letterpaper]{article} 
\usepackage{aaai21}  
\usepackage{times}  
\usepackage{helvet} 
\usepackage{courier}  
\usepackage[hyphens]{url}  
\usepackage{graphicx} 
\usepackage{natbib}  
\usepackage[noend]{algpseudocode}
\usepackage{algorithmicx,algorithm}
\usepackage{booktabs}
\usepackage{multirow}
\usepackage{amsmath}
\usepackage{subfigure}
\usepackage{caption} 
\usepackage{color,xcolor}
\usepackage{amsfonts}
\usepackage{mathtools}
\usepackage[switch]{lineno}

\definecolor{reds}{RGB}{184,84,80}
\definecolor{blues}{RGB}{16,115,158}

\title{Contrastive Triple Extraction with Generative Transformer}

\author {
        Hongbin Ye\textsuperscript{\rm 1,2 \footnote{Equal contribution and shared co-first authorship.}},
        Ningyu Zhang\textsuperscript{\rm 1,2*\dag},
        Shumin Deng\textsuperscript{\rm 1,2},
        Mosha Chen\textsuperscript{\rm 3},
        Chuanqi Tan\textsuperscript{\rm 3}, \\
        Fei Huang\textsuperscript{\rm 3}
        Huajun Chen\textsuperscript{\rm 1,2 \footnote{Corresponding author.}}
        \\
}
\affiliations {
    \textsuperscript{\rm 1} 
     Zhejiang University\\
    \textsuperscript{\rm 2} AZFT Joint Lab for Knowledge Engine
    \textsuperscript{\rm 3} Alibaba Group \\
    {\tt \{yehb,zhangningyu,231sm,huajunsir\}@zju.edu.cn,  \{chenmosha.cms,chuanqi.tcq,f.huang\}@alibaba-inc.com }
}

\begin{document}
\maketitle

\begin{abstract}
Triple extraction is an essential task in information extraction for natural language processing and knowledge graph construction. In this paper, we revisit the end-to-end triple extraction task for sequence generation. Since generative triple extraction may struggle to capture long-term dependencies and generate unfaithful triples, we introduce a novel model, contrastive triple extraction with a generative transformer. Specifically, we introduce a single shared transformer module for encoder-decoder-based generation. To generate faithful results, we propose a novel triplet contrastive training object.  Moreover, we introduce two mechanisms to further improve model performance (i.e., batch-wise dynamic attention-masking and triple-wise calibration).  Experimental results on three datasets (i.e., NYT, WebNLG, and MIE) show that our approach achieves better performance than that of baselines. 
\end{abstract}
\section{Introduction}
Triple extraction is an essential information extraction task for natural language processing (NLP) and knowledge graph (KG), which is used to detect pairs of entities and their relations from unstructured text. Consider this sentence: ‘‘\emph{Paris is known as the romantic capital of France}.'' From this, an ideal triple extraction would comprise {\it $\langle$Paris, Capital\_of, France$\rangle$}, in which {\it Capital\_of} is the relation of {\it Paris} and {\it France}.

\begin{table}[h] \centering
\begin{tabular}{l|p{6.5cm}}
 \toprule
    Input&  The \emph{United States} President \emph{Trump} was raised in the borough of \emph{Queens} in \emph{New York City}, and lived there until age 13.  \\
    \midrule
    Output& Trump$\rightarrow$president$\rightarrow$of$\rightarrow$United$\rightarrow$ States$\rightarrow$[S2S\_SEQ]$\rightarrow$Trump$\rightarrow$born$\rightarrow$in$\rightarrow$ Queens$\rightarrow$[S2S\_SEQ]$\rightarrow$Trump$\rightarrow$live$\rightarrow$in$\rightarrow$Queens \\
    \midrule
    \multirow{3}{*}{Gold}& (Trump, president\_of, United States)\\ &(Trump, born\_in, Queens) \\ &(Trump, live\_in, Queens) \\
    \midrule
    \multirow{3}{*}{Negative}& (Trump, president\_of, Queens)\\ &(Trump, born\_in, 13) \\ &(Trump, live\_in, 13) \\
 \bottomrule
\end{tabular}
\caption{Contrastive triple extraction  as sequence generation. We encourage the model to generate gold triples and does not generate negative ones.}
\label{nyt}
\end{table}

Researchers have proposed pipeline approaches in the past \cite{Lample2016NeuralAF,zeng2015distant} in which they typically deconstructed the triple extraction problem into two separate tasks: named-entity recognition (NER) (used to extract entities) and relation classification. Thus, they first recognized the entities; then, they predicted their relationships. Unfortunately, this and similar pipeline approaches suffer drawbacks \cite{roth2007global} in that they omit the evident correlations between entity recognition and relation extraction tasks, resulting in error propagation. 

Recently, several neural-network-based models \cite{zeng2018extracting} have been proposed to jointly extract entities and relations from sentences. These models use a parameter-sharing mechanism to extract entities and relations from the same network. Apart from those approaches, \citet{zeng-etal-2018-extracting} proposed a recurrent neural-network-based encoder-decoder model (i.e., CopyRE) to extract triples with overlapping entities. Such end-to-end generative triple extraction not only directly obtain the triples and mitigate the error propagation issue, but also enable the generation of out of domain entities and relations in a T5-style  \cite{raffel2019exploring} (text-to-text).  Besides, \citet{zeng2020copymtl} proposed a multi-task learning framework equipped with a copy mechanism (i.e., CopyMTL) to allow the prediction of multi-token entities. \citet{nayak2019effective} introduced a representation scheme for triples and a pointer-network-based decoding approach, which further improved the performance of CopyRE.

Encoder-decoder models are powerful tools that have seen success in many NLP tasks, including machine translation \cite{cho2014learning}, 
and open information extraction \cite{zhang2017selective}. Although significant progress has been achieved, there remain two key problems with the existing methods. \textbf{First}, owing to the intrinsic shortfalls of recurrent neural networks (RNN), they cannot capture long-term dependencies, which results in the loss of important information otherwise reflected in the sentence. Such a drawback prevents the model from being applied to longer texts. \textbf{Second}, there is a scarcity of work that has focused on generating faithful triples. As a previous study \cite{zhu2020boosting} indicated, a sequence-to-sequence architecture can generate unfaithful sequences that create contradictions of meaning. For example, given the sentence ``The \emph{United States} President \emph{Trump} was raised in the borough of \emph{Queens} in \emph{New York City}, and lived there until age 13,’’  the model could generate the fact ``(Trump, born\_in, Queens).’’ Although logically true, we cannot find direct evidence from the given sentence to support it. 

To address these issues, we introduce a framework of \textbf{C}ontrastive triple extraction with \textbf{G}enerative \textbf{T}ransformer (CGT), which is a single shared transformer module with a triplet contrastive object that supports encoder-decoder generation. To begin with, we concatenate the input sequence with the target sequence using a separator token and leverage partial causal masking \cite{du2020document} to distinguish the encoder-decoder representations. Our model requires no additional parameters beyond those of the pre-trained model. Then, we introduce a novel triplet contrastive learning object, which utilizes ground-truth triples as positive instances and leverages random token sampling to construct corrupt triples as negative instances. To jointly optimize the triple generation and contrastive object, we introduce a batch-wise dynamic attention-masking mechanism, which allows us to dynamically choose different objects and jointly optimize tasks. Lastly, we introduce a novel triple-wise calibrating algorithm to filter out any remaining false triples in the inference stage. 

The contributions of this work are as follows: 
\begin{itemize}
    \item We revisit triple extraction as a sequence generation task and introduce a novel CGT model. In light of the added extraction capability, CGT requires no additional parameters beyond those found in the pre-trained language model.
    \item We introduce two mechanisms to further improve model performance (i.e., batch-wise dynamic attention-masking and triple-wise calibration). The first enables joint optimization of different objects, and the second ensures faithful inference.
    \item We evaluate CGT on three benchmark datasets. Our model empirically outperforms other substantially strong baseline models. We also demonstrate that CGT is better than existing triple extraction approaches at capturing long-term dependencies, thus, achieving better performance with long sentences.
\end{itemize}
 
\section{Related Work}
\subsection{Triple Extraction}
Two main methods have been proposed for triple extraction: pipeline \cite{Nadeau2007ASO,bunescu2005shortest,lin2016neural,lin2017neural,li-etal-2020-logic,wang2020finding} and joint learning \cite{miwa2016end,katiyar2017going,cao-etal-2017-bridge,zhang-etal-2020-openue,dai2019joint}. A pipeline method first extracts entities, then it identifies their relations \cite{hendrickx2019semeval,zeng2015distant}. Although pipeline models have achieved great progress \cite{zhang2018capsule,he2018see,zhang2019long,zhang2020relation}, they introduce an error propagation problem \cite{li2014incremental}, which does harm to the overall performance.

Because joint learning can implicitly model correlations between tasks, many approaches have been proposed. \citet{bekoulis2018joint} formulated the triple extraction task as a multi-head selection problem. \citet{takanobu2019hierarchical} proposed a hierarchical reinforcement-learning framework for triple extraction. \citet{chen2019mrmep} utilized triplet attention to exploit connections between the relation and its corresponding entity pairs. \citet{2019Joint} introduced a position-attention mechanism to produce different tag sequences for triple extraction.\citet{wei2019joint} revisited the relational triple extraction task and proposed a novel cascade binary-tagging framework. Apart from those approaches, \citet{zeng2018extracting} proposed CopyRE, a joint model based on a copy mechanism, which converted the joint extraction task into a triplet-generation task. Other researchers  introduced multiple strategies, such as multi-task learning \cite{zeng2020copymtl} and one-word generation \cite{nayak2019effective} to improve CopyRE. For the first time, we utilize the transformer as an encoder-decoder architecture to extract triples from sentences.

\subsection{Natural Language Generation} 
Natural language generation has been intensively studied in the recent literature. Most models employed an encoder-decoder architecture (i.e., seq2seq) using RNNs \cite{schuster1997bidirectional,zhang-etal-2020-summarizing,krause2020gedi}. Recently, owing to the powerful representation ability of transformers, several researchers have introduced transformer-based natural language generation methods. \citet{gu2019levenshtein} developed the Levenshtein transformer, a new partially autoregressive model, which is devised for a more flexible and amenable sequence generation. \citet{chen2020distilling} present a novel approach, Conditional Masked Language Modeling (C-MLM), to enable the finetuning of BERT \cite{devlin2018bert} on target generation tasks. \citet{dong2019unified} proposed a new unified pre-trained language model with different masking strategies, which can be used for both language understanding and generation. \citet{du2020document} proposed a generative transformer-based encoder-decoder framework for document-level information extraction.

Since the generation procedure was unconditional, it was non-trivial to judge the faithfulness of the generated sequence. \citet{zhang2019optimizing} approached the factual correctness problem in the medical domain, where the space of facts was limited and could be depicted with a descriptor vector. \citet{cao2017faithful} extracted relational information from an article and mapped it to a sequence as input to the encoder. The decoder then attended to both article tokens and their relations. \citet{gunel2020mind} employed an entity-aware transformer structure to boost the factual correctness of abstractive summarization, where the entities came from the Wikidata knowledge graph. By comparison, our model utilizes contrastive learning to encourage the model to implicitly generate faithful triples.

\begin{figure*}[h]
\centering
  \includegraphics[width=1\textwidth]{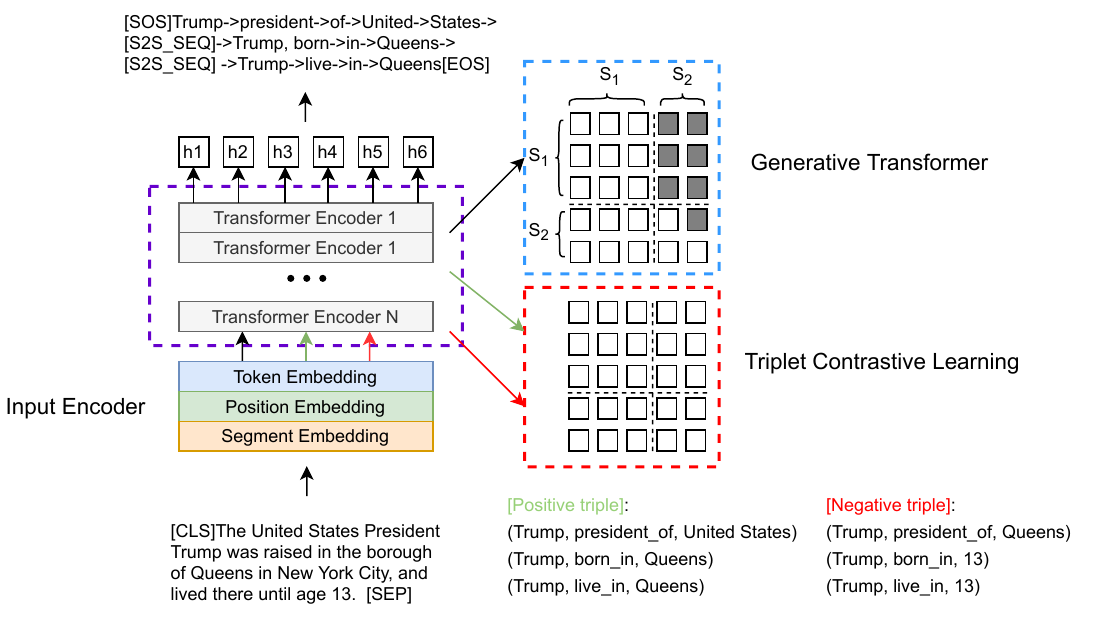}
\caption{The architecture of \textbf{C}ontrastive triple extraction with \textbf{G}enerative \textbf{T}ransformer (CGT).  The top-right component refers to the generative transformer, and the bottom-right component represents triplet contrastive learning. Those two parts are optimized jointly. The left is the input encoder (best viewed in color).}
\label{arc}
\end{figure*}

\section{Overview}

\subsection{Preliminary}
We treat triple extraction as a sequence-to-sequence task to better model the cross dependencies between entities and relations. We define the input text and output triples as source and target sequence. As shown in Figure \ref{arc}, the source sequence simply consists of the tokens of the input sentence like ``[CLS] The United States President Trump was raised in the borough of Queens ...[SEP]". We concatenate the triples for each entity/relation separated by a special token token [S2S\_SEQ] as the target sequence. We also add the beginning ([SOS]) and end ([EOS]) tokens for each target sequence as:
\begin{center}
$  \begin{array}{c}
\mathrm{[SOS]}
h^{(1)}, r^{(1)}, t^{(1)}\ldots[\mathrm{S2S\_SEQ}] \\
h^{(2)}, r^{(2)}, t^{(2)}\ldots[\mathrm{S2S\_SEQ}] \\
h^{(3)}, r^{(3)}, t^{(3)}\ldots[\mathrm{S2S\_SEQ}] \\
...\\
h^{(N)}, r^{(N)}, t^{(N)}\ldots[\mathrm{EOS}], \\
\end{array}$
\end{center}
 
where $h^{i}$, $r^{i}$, and $t^{i}$ refer to the $i$-th generated head entity, relation, and tail entity. 
\subsection{Framework}

We denote the sequence of input source tokens as $x_0$, $x_1$, ..., $x_m$ and the sequence of target tokens as $y_0$, $y_1$, ..., $y_n$. Note that the generated tokens contain all extracted triples. Our model CGT consists of three components, as follows:

\textbf{Input Encoder.} We utilize the input representation which  is the same as BERT \cite{devlin2018bert} and tokenize texts by WordPiece \cite{yonghui2016bridging}. We compute the representation by summing the corresponding token embedding, position embedding, and segment embedding. 

\textbf{Generative Transformer.} We use partial causal masking to distinguish the encoder-decoder representations. For inference, we leverage the beam search \cite{wiseman2016sequence} to generate multiple triples.  

\textbf{Triplet Contrastive Learning.} We introduce a triplet contrastive object to enhance the faithfulness of generated triples. We introduce a batch-wise dynamic attention masking mechanism for joint optimization. We also provide a triple-wise calibrating algorithm for the faithful triple generation. 

\section{Our Model}
\subsection{Input Encoder}
Given the input text $x$, we add a special start-of-sequence token [SOS] at the beginning of the target input. We use the representation of the whole input for the output vector.  Furthermore, we append a special token, namely, end-of-sequence [EOS], to the end of each output sequence. The [EOS] token is used as a special token to terminate the decoding process for the triple generation.  

The input representation is the same as the one used for BERT \cite{devlin2018bert}. We tokenize the text to subword units using WordPiece \cite{yonghui2016bridging}. For example, the word, ``forecasted,’’ is split into ``forecast’’ and ``\#\#ed,’’ where ``\#\#’’ refers to the pieces belong to one word. We compute each input token vector representation by summing the corresponding token embedding, position embedding, and segment embedding.  


 
\subsection{Generative Transformer}

We utilize a  transformer architecture as a backbone to encode contextual features which is consist of stacked self-attention layers. In this paper, we use a 12-layer transformer architecture as a single shared transformer module for encoder-decoder-based generation. Having the input vectors, $\left\{\mathbf{s}_{i}\right\}_{i=1}^{|L|}$, we firstly feed them into $\mathbf{H}^{0}=\left[\mathbf{s}_{1}, \cdots, \mathbf{s}_{|L|}\right]$. Then, we use the transformer to encode the input:
\begin{equation}
    \mathbf{H}^{l}=\operatorname{Transformer}_{l}\left(\mathbf{H}^{l-1}\right).
\end{equation} 

There are multiple self-attention heads in each transformer block which are used to aggregate the output vectors of the previous layer. We compute the output of a self-attention head, $A_l$, in the $l$-th transformer layer as follows:
\begin{equation}
    \mathbf{Q}_{l}=\mathbf{H}^{l-1} \mathbf{W}_{l}^{Q}, \quad \mathbf{K}_{l}=\mathbf{H}^{l-1} \mathbf{W}_{l}^{K}.
\end{equation}

\begin{equation}
\mathbf{M}_{i j}=\left\{\begin{array}{ll}0, & \text { allow to attend } \\ -\infty, & \text { prevent from attending }\end{array}\right.
\end{equation}

\begin{equation}
    \textbf{$\mathbf{A}_{l}=\operatorname{softmax}\left(\frac{\mathbf{Q}_{l} \mathbf{K}^{\top}}{\sqrt{d_{k}}}+\mathbf{M}\right)\left(\mathbf{H}^{l-1} \mathbf{V}_{l}\right)$},
\end{equation}

where $\mathbf{Q}_{l}, \mathbf{K}_{l}, \mathbf{V}_{l} \in \mathbb{R}^{d_{h} \times d_{k}}$ are  matrices which are the  projection of the the previous layer’s output.
The mask matrix, $\mathbf{M} \in \mathbb{R}^{|L| \times|L|}$, is aimed to control the context that can be attended by the token. Specifically, we leverage different mask matrices, $\mathbf{M}$,  when computing its contextualized representation. 
As illustrated by the examples in Figure \ref{arc}, for triple generation, we leverage partial causal masking, in which  the upper right part is set to $-\infty$ to block attention from the source segment to the target segment; the left part of $\mathbf{M}$ is set to all $0$s which indicates that the tokens is able to attend to the first segment. We utilize cross-entropy $\mathbf{loss}_{generation}$ to optimize the triple generation procedure. We also utilize masking strategies in which the elements of the mask matrix are all $0$s for triplet contrastive learning. Details are provided in the next section. Formally, the generative transformer obtain contextualized representations and optimize the following object:
\begin{equation}\begin{split}
    &\hat{\mathbf{x}}_{0}, \hat{\mathbf{x}}_{1}, \ldots, \hat{\mathbf{x}}_{m}, \hat{\mathbf{y}}_{0} \ldots, \hat{\mathbf{y}}_{n} \\
    &=\operatorname{Transformer}\left(\mathbf{x}_{0}, \mathbf{x}_{1}, \ldots, \mathbf{x}_{m}, \mathbf{y}_{0}, \ldots, \mathbf{y}_{n}\right)
    \end{split}
\end{equation}

\begin{equation}
\mathbf{loss}_{generation} = -\sum (\sum_{1}^m \mathbf{x}_{i} log(\hat{\mathbf{x}}_{i}) + \sum_{1}^n  \mathbf{y}_{i} log(\hat{\mathbf{y}}_{i}))
\end{equation}

\subsection{Triplet Contrastive Learning}
The previous generation-based approach usually neglects the fact that triple should be faithful and consistent with the input sentence. For example, given the instance ``Obama was born in Honolulu," we should engorge the model to generate triples like ``(Obama, was\_born, Honolulu)" rather than ``(Obama, live\_in, Honolulu)," though the latter may be correct but cannot be induced from the given sentence. Motivated by this, we introduce a triplet contrastive learning to enhance the faithfulness of generated triples.

To be specific, we leverage the triple contrastive learning as binary classification with all $0$s masking. We use gold triples as positive instances and generate corrupt triples by replacing one entity with random tokens in the instances. We use those corrupt triples as negative instances. We concatenate the input sentence with only one triple as $x_0$, $x_1$, ..., $x_m [SEP], h^{i},r^{i},t^{i}$ and feed it into the input encoder. We utilize the representation of [CLS] with an MLP layer to compute classification logits $z$. We utilize cross-entropy for optimization with $\mathbf{loss}_{contrastive}$:

\begin{equation}
\mathbf{loss}_{contrastive} = -\sum (\mathbf{z}^{+}_{i} log(\hat{\mathbf{z}}^{+}_{i}) + (1-\mathbf{z}^{-}_{i}) log((1-\hat{\mathbf{z}}^{-}_{i})))
\end{equation}

where $\hat{\mathbf{z}}^{+}_{i}$ and $\hat{\mathbf{z}}^{-}_{i}$ are the positive and negative logits, respectively. Formally, the triplet contrastive learning algorithm for triple extraction is as follows:

\begin{algorithm}[th]
\begin{algorithmic}[1]
\caption{Triplet Contrastive Learning} 
 \State \textbf{Require:} Train instances $X={x_1,...,x_N}$, labels $Y={y_1,...,y_N}$,  batch size $k$, $POS = \Phi$, $NEG = \Phi$, temperature $t$
 \While{$i$ $\le$ $N/k$ }
   \State batch = $[(x,y)_1,..,(x,y)_k]$
   \For{$(x,y)_j$ in batch}
    \State POS = decompose\_triple($y_j$)
    \For{$pos$ in POS}
    \State neg = random\_permute\_entity(pos)
    \State l\_pos = Contrastive\_Classify(x,pos)
    \State l\_neg = Contrastive\_Classify(x,neg)
    \State z = cat([l\_pos, l\_neg], dim=1)
    \State labels = zeros(2) 
    \State loss = CrossEntropyLoss(z/t, labels)
    \State loss.backward()
    \State update(Contrastive\_Classifier.params)
     \EndFor
    \EndFor
   \EndWhile
   \State return DataLoader
\label{batch} 
\end{algorithmic}
\end{algorithm}

\subsection{Training and Inference Details}

During the training stage, the entities and relations are all tokens from the vocabulary, whereas [S2S\_SEQ], [SOS], and [EOS] are all unused tokens (e.g., [unused1]). We split the entity and relation label mentions with different tokens during the data preprocessing procedure, meaning that the entity and relation may contain multiple tokens. 

Note that triplet contrastive learning and triple generation are two different tasks, and optimizing them jointly is non-trivial, owing to the leakage of generated labels. For example, if we optimize generation and contrastive learning with the same instance, the model can see all of the tokens because of the all $0$s masking. To address this issue, we introduce batch-wise dynamic attention masking. With this, we sample instances from a Bernoulli distribution as generation instances, and the rest is sampled as contrastive learning sentences. Formally, the algorithm is as follows:

\begin{algorithm}[th]
\begin{algorithmic}[1]
\caption{Batch-wise  Dynamic  Attention Masking} 
 \State \textbf{Require:} Train instances $X={x_1,...,x_N}$, labels $Y={y_1,...,y_N}$, negative instances $Y^{\prime}$,  batch size $k$ sampling ratio $\gamma$
 \While{$i$ $\le$ $N/k$ }
   \State old\_batch = $[(x,y,y^{\prime})_1,..,(x,y,y^{\prime})_k]$
   \For{$(x,y,y^{\prime})_j$ in old\_batch}
    \State condition = Bernoulli($\gamma$)
    \If{condition $==$ 1}
    \State    instance = Partial\_Causal\_Mask($(x,y,y^{\prime})_j$)
     \Else
     \State   instance = All\_Zero\_Mask($(x,y,y^{\prime})_j$)
    \EndIf
    \State batch $\leftarrow$ batch $\cap$  instance
    \EndFor
    \State DataLoader  $\leftarrow$ DataLoader $\cap$  batch
    \State batch = $\Phi$
   \EndWhile
   \Return DataLoader
\label{batch} 
\end{algorithmic}
\end{algorithm}
The overall optimization object is as follows:
\begin{equation}
    \mathbf{loss} = \mathbf{loss}_{generative} + \alpha \mathbf{loss}_{contrastive}
\end{equation}
where $\alpha$ is the hyperparameter to balance different objects. 

During the inference stage, we first generate triplet sequences via beam search \cite{wiseman2016sequence}. Then, we introduce a triple-wise calibrating algorithm to filter-out unfaithful triples. We calculate the matching score with the contrastive classifier and filter out those triples with the $match\_score < \theta$. Besides, we also leverage heuristic rules to generate reasonable triples such as the relation should be followed by the head entities. 
\section{Experiment}
\subsection{Dataset}

We conducted experiments on three benchmark datasets: New York Times (NYT), WebNLG\footnote{\url{https://github.com/weizhepei/CasRel}}, and MIE\footnote{\url{https://github. com/nlpir2020/MIE-ACL-2020}}. The NYT dataset is produced using a distant supervision method and is widely used for triplet extraction \cite{riedel2010modeling}.  It contains 56,195 sentences for training, 5,000 sentences for validation, and 5,000 sentences for test. The WebNLG dataset \cite{gardent2017creating} was used for natural language generation, but was later used for triplet extraction \cite{zeng2018extracting}. It consists of 5,019/500/703 instances for training, validation, and testing, respectively. MIE \cite{zhang2020mie} is a large-scale Chinese dialogue information extraction dataset for the medical domain. It contains 800 instances for training, 160 instances for validation, and 160 instances for testing. We used the original dataset splitting for NYT, WebNLG, and MIE. Detailed statistics of the three datasets are shown in Table \ref{dataset}
\begin{table}[!htbp]
 
\centering
{
\begin{tabular}{c|ccc}

\toprule
Dataset&NYT&WebNLG&MIE\\
\midrule
Domain&News&Web&Medical\\
Relation&24&246&343\\
Triplets&104,518&12,863&18,212\\
\bottomrule
\end{tabular}
}
\caption{Statistics of four datasets in the domain, the number of relation types, and the triple number.}
\label{dataset}
\end{table}

\subsection{Settings}
We utilized \emph{UniLM-base-uncased} for both English\footnote{\url{https://github.com/microsoft/unilm}} and Chinese\footnote{\url{https://github.com/YunwenTechnology/Unilm}} datasets. We utilized Pytorch  to implement our CGT model and conducted experiments using four Nvidia 1080-Ti graphical processing units. We employed Adam~\cite{kingma2014adam} as the optimizer. The initial learning rate was set to 2e-5, and we reduced the rate by 20\% at every eight epochs. The batch size was 64 for English and 32 for Chinese, and the total number of epochs was 50 for all datasets. The beam size was set to 4, $\alpha$ was set to 0.1,  $\gamma$ was set to 0.2, and $\theta$ was set to 0.6. We carefully tuned the hypermeters on the valid set (Detailed search space in supplementary materials). 

\subsection{Baselines and Evaluation Metrics}
We compared the performance of CGT with various baseline models and evaluated the performance with precision, recall, and F1 score. CGT(Random) and CGT(UniLM) refer to the model initialized randomly, and the model initialized with UniLM, respectively.

\textbf{Generative Baseline Models:}

\begin{table*}[h] \centering
\begin{tabular}{ll|ccc|ccc}
 \toprule
 \multicolumn{2}{c|}{\multirow{2}{*}{\textbf{Model}}}& \multicolumn{3}{c|}{\textbf{NYT}} & \multicolumn{3}{c}{\textbf{WebNLG}} \\
\cline{3-8}
&& P & R & F & P & R & F  \\
 \midrule
   \multirow{4}{*}{Extractive}    &  Tagging \cite{DBLP:journals/corr/ZhengWBHZX17}& 61.5 &41.4 &49.5 &- & - &- \\
       & HRL\cite{takanobu2019hierarchical} & 71.4 & 58.6 & 64.4 & 53.8& 53.8 &53.8  \\
      & MrMep \cite{chen2019mrmep} &77.9  &76.6  &77.1  &69.4 &77.0  &73.0  \\
      & CasRel \cite{wei2020novel}  &89.7 &89.5 &89.6 &93.4  &90.1  &91.8  \\
       \midrule
   \multirow{3}{*}{Generative}   &  CopyRE \cite{zeng2018extracting} & 61.0& 56.6& 58.7 &37.7& 36.4& 37.1 \\
       & PNDec \cite{nayak2019effective} & 80.6& 77.3& 78.9 &38.1& 36.9& 37.5 \\
    &    CopyMTL  \cite{zeng2020copymtl}   &75.7  &68.7  &72.0  &58.0 &54.9 &56.4  \\
\midrule
\multirow{3}{*}{Ours}& CGT(Random) & 90.8  & 77.7  & 83.7 &87.6 &70.5  &78.1 \\ 
& CGT(UniLM) & \textbf{94.7*} & 84.2 & 89.1  & \textbf{92.9*}   & 75.6   & 83.4  \\
& w/o contrastive&87.3 &81.5 &84.3 &94.6 &70.5 &80.8\\
 \bottomrule
\end{tabular}
\caption{Main results of NYT and WebNLG. The top section refers to the extractive models, the middle section indicates the generative approaches, the bottom is our model with different settings. * indicates $p_{value} < 0.01$ for a paired t-test evaluation.}
\label{nyt}
\end{table*}

\textbf{CopyRE} \cite{zeng2018extracting} is a Seq2Seq learning framework having a copy mechanism wherein multiple decoders are applied to generate triples to handle overlapping relations.

\textbf{PNDec} \cite{nayak2019effective} provides two novel approaches using encoder-decoder architecture for triples having multiple tokens.

\textbf{CopyMTL} \cite{zeng2020copymtl} proposes a multitask learning framework used to complete the entities. 
 
\textbf{Extractive Baselines:}

\textbf{Tagging} \cite{zheng2017joint} is an end-to-end method that uses a novel tagging scheme. 

\textbf{HRL} \cite{takanobu2019hierarchical} addresses relation extractions by regarding related entities as the arguments of the relation via hierarchical reinforcement learning. 

\textbf{MrMep} \cite{chen2019mrmep} is an approach that utilizes triplet attention to exploit connections between relations and their corresponding entity pairs.

\textbf{CasRel} \cite{wei2019joint} is an approach that models relations as functions, which map subjects to objects in a sentence.

\textbf{Bi-LSTM} \cite{zhang2020mie} is a baseline approach that utilizes a bi-directional long-short term memory network for information extraction. 

\textbf{MIE-multi} \cite{zhang2020mie} is another baseline model that uses a max-pooling operation to obtain the final score, considering the turn-interaction. 

\subsection{Main Results}

From Table \ref{nyt}, we observe that our approach achieved significant improvements compared with all generation-based baseline models for both NYT and WebNLG datasets. Our CGT model had a relative \textbf{10.2} F1 score improvement on NYT compared with PNDec, and a relative \textbf{27.0} F1 score improvement on NYT compared with CopyMTL, illustrating the power of our proposed model. Our approach also obtained comparable results compared with extractive models, such as CasRel. Note that the search space of the generative model was much larger than the extractive ones, which indicates that the generative model was challenging to optimize than extractive approaches. In contrast, generative methods can generate triples beyond the entity and relation domain, which is promising for the open domain setting. The empirical results reveal that the generative approach could obtain comparable performance with extractive models, motivating future research directions. 

 \begin{table}[!htbp]
\centering
\begin{tabular}{c|ccc}
\toprule
Model& P&R&F1 \\
\midrule
Bi-LSTM &53.13 &49.46& 50.69 \\
 MIE-multi &70.24 & 64.96 & 66.40 \\
 \midrule
 CGT(random)&70.75  &66.96  &68.80 \\
 CGT(UniLM)& \textbf{80.53}&\textbf{78.83} &\textbf{79.42} \\
\bottomrule
\end{tabular}
\caption{Main results on the MIE dataset. }
  \label{mie}
\end{table}

From Table \ref{mie}, we observe that our approach achieved significant improvements (relative \textbf{13.02} F1 score) compared with all baselines on the MIE dataset. MIE is a dialogue-based information-extraction dataset that is challenging to optimize. Thus, we argue that our CGT can implicitly model the relations among entities, boosting performance. 

\subsection{Ablation Study}

We conducted ablation studies further to demonstrate the efficacy of different strategies in our model. From Table \ref{nyt}, we notice that the performance decayed without contrastive object, which illustrates that triplet contrastive learning can enhance the faithfulness of generated triples, thus boosting the performance. We also observe that our approach with random initialization CGT(Random) achieves significantly better performance than generative baselines on all three benchmark datasets, which further indicates that our improvements are not only from the pre-trained language model but also the model architecture itself. 

\begin{figure*}[!htbp]
\centering
\subfigure[NYT] { 
  \includegraphics[scale=0.191]{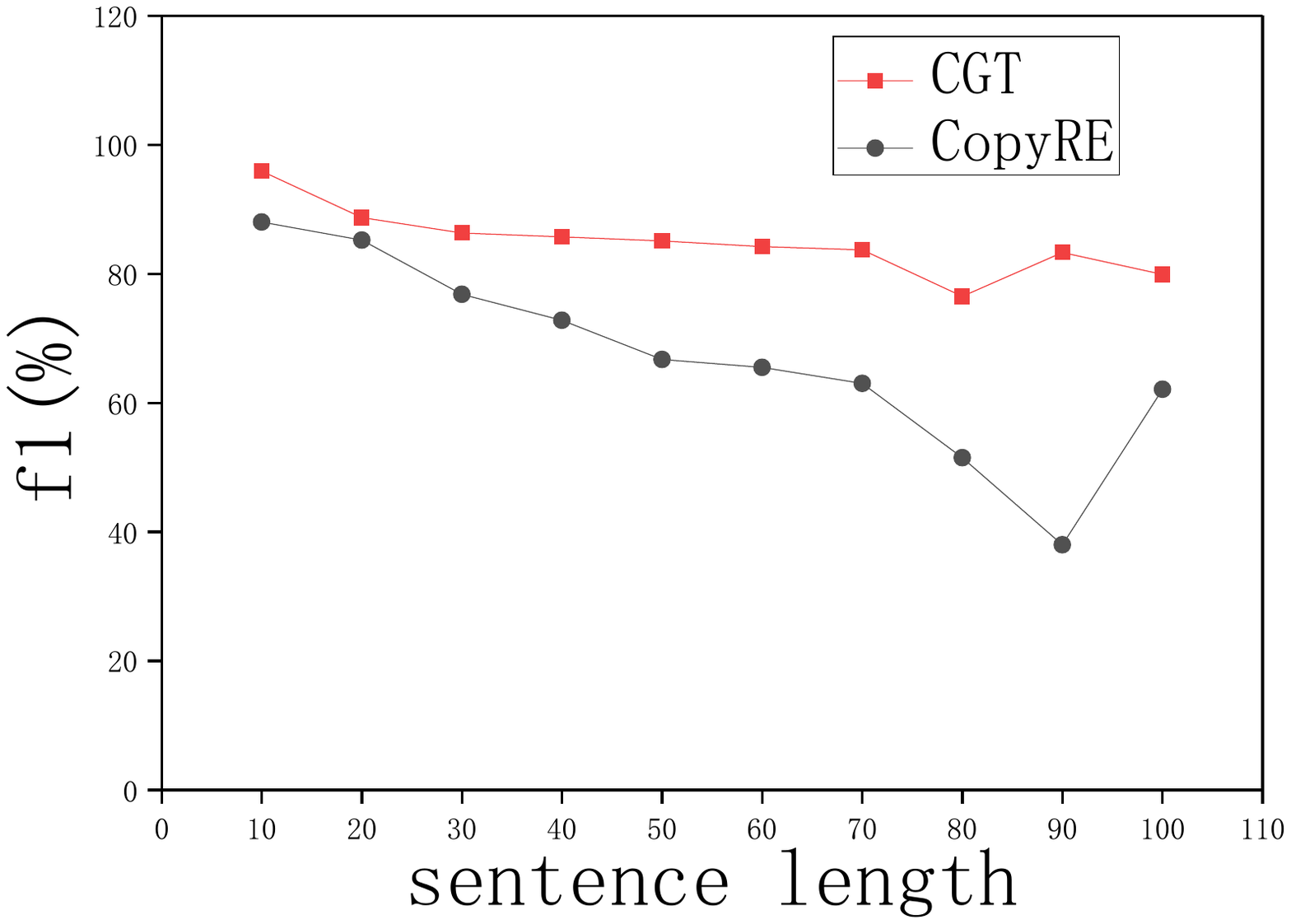}\label{f1}
}
\subfigure[WebNLG] { 
\includegraphics[scale=0.191]{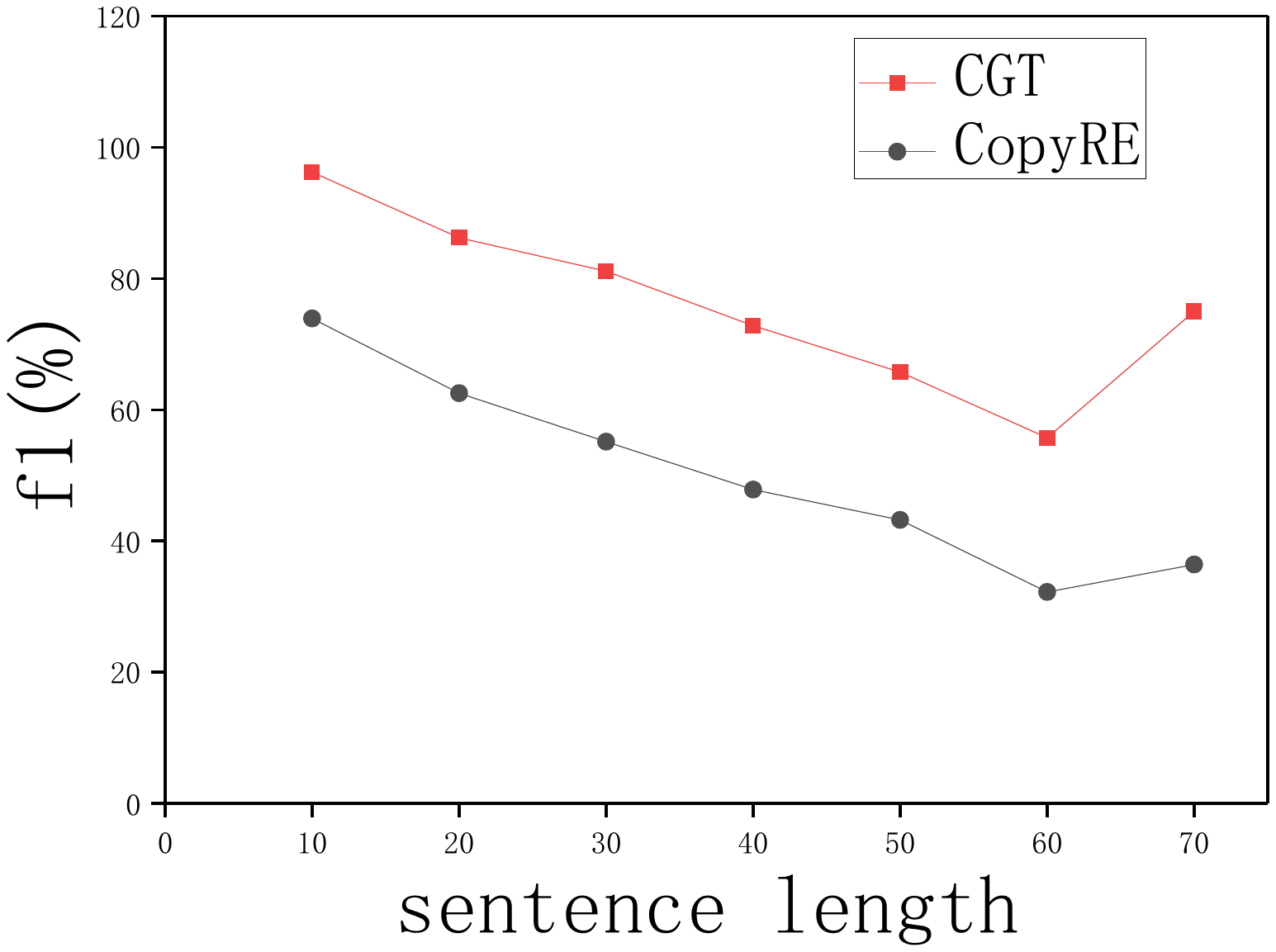}\label{f2}
}
\subfigure[MIE] { 
\includegraphics[scale=0.191]{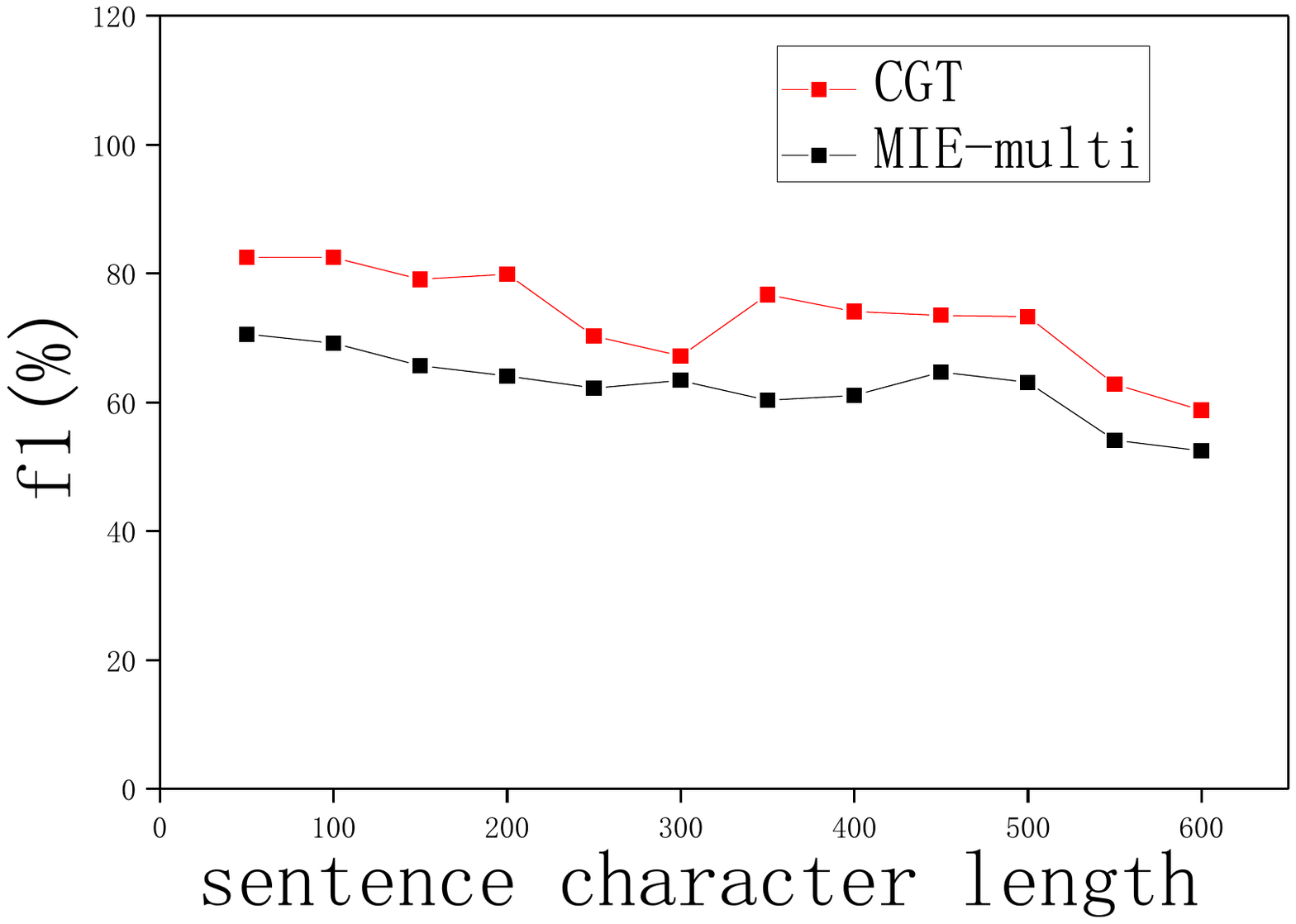}\label{f3}
}
\caption{Model performance \#\emph{sentence length}.}
\label{len}
\end{figure*}

\begin{table*}[h]
  \centering 
  \resizebox{\textwidth}{!}{
\begin{tabular}{l}
  \toprule
  {\bf Instance}\\
  \midrule
  \underline{instance \#1} \ \quad
  Batchoy is originates from the Philippines and served as a soup.Its main ingredients are noodles,\\ pork organs, vegetables, chicken, shrimp and beef.\\

  generated triple: 
  {\it $\langle${\color{reds}{Batchoy}}, location, {\color{blues}{Philippines}}$\rangle$} \\

  ground truth: \ \quad
  {\it $\langle${\color{reds}{Batchoy}}, country, {\color{blues}{Philippines}}$\rangle$} \vspace{1 ex}\\
   
  \underline{instance \#2} \;
  Alan Shepard was a crew member of NASA operated Apollo 14 who died in California which \\is represented by Dianne Feinstein.\\

  generated triple: 
  {\it $\langle${\color{reds}{Shepard}}, deathPlace, 
  {\color{blues}{California}}$\rangle$} \\

  ground truth: \ \quad
  {\it $\langle${\color{reds}{Allan Shepard}}, deathPlace, 
  {\color{blues}{California}}$\rangle$} \\
  \vspace{1 ex} \\
  
  \underline{instance \#3} \;
   Saranac Lake, which is served by Adirondack Regional Airport, is part of Harrietstown,  Essex \\ County, New York, US. \\

  generated triple: 
  {\it $\langle${\color{reds}{Airport}}, cityServed, {\color{blues}{New York}}$\rangle$} \\

  ground truth: \ \quad
 {\it $\langle${\color{reds}{Airport}}, cityServed, {\color{blues}{York}}$\rangle$} \\

  \bottomrule
  \end{tabular}
  }
\caption{Error anslysis.}
\label{errorAnylysis}
\end{table*}

\subsection{Analysis}
To better analyze the performance of our proposed CGT model, we conducted a detailed analysis and attempted to answer the questions of whether CGT can capture long-term dependence or not.
Intrusively, transformers having self-attention can better capture long-term dependencies than RNNs. To investigate this issue, we evaluated the instances at different lengths. From Figure \ref{len}, we notice that all models have a performance decay when the sentence length increases, which indicates that the sequence generation is challenging when the input sentence is long. We observe that our approach could obtain better performance than that of CopyRE when the sentence length increased. When the sentence was longer than 60, CopyRE archived worse performance, while CGT performed relatively better. This demonstrates that the proposed approach can capture long-term dependencies, compared with RNN-based approaches.
 
\subsection{Error Analysis}
To further analyze the drawbacks of our approach and promote future works of triple extraction, we select instances and conduct error analysis. We random select incorrect instances and classify them into three categories bellows, 
as shown in Table \ref{errorAnylysis}:

\textbf{Distract Context.} As instance \#1 shows, we observe that our approach may fail to those ambiguous contexts that may be expressed in a similar context but differ only in the fine-grained type of entities. We argue that this may be caused by the unbalanced learning problems that models tend to judge the sentence with similar context to high-frequency relations. 

 \textbf{Wrong Boundaries.} As instance \#2 shows, generated triples had incorrect boundaries, which indicates the difficulty of entity recognition during triple extraction. We argue that since our approach is an end-to-end generation method, it is challenging to capture fine-grained entity boundaries without sequence token information. 
 
 \textbf{Wrong Triples.} As instance \#3 shows, many generated triples had entities that did not exist in the gold-standard set. Generally, this occurs with sentences having multiple triples. The WebNLG datasets are noisy, and several of its cases produced incorrect results. We leave this for future works with more suitable benchmarks.  
 
\section{Conclusion and Future Work}
In this paper, we revisited triple extraction as a sequence generation task, which jointly extracts entities and relations. To address the long-term dependence and faithfulness issues, we proposed a novel CGT model to generate faithful triples. To the best of our knowledge, we are the first to integrate sequence generation with contrastive learning for information extraction, which may inspire future research directions and motivate new ideas. Experimental results on three datasets demonstrated the efficacy of our approach. In the future, we will utilize stronger transformer architectures, such as Longformer \cite{beltagy2020longformer} to generate relational knowledge from documents. We will also delve into injection ontology knowledge using condition generation methods. It will also be useful to apply our approach to other scenarios, such as event extractions. 

\section{Acknowledgments}
We  want to express gratitude to the anonymous reviewers for their hard work and kind comments. We thank Ning Ding for helpful discussions and feedback on this paper. This work is funded by 2018YFB1402800/NSFC91846204/NSFCU19B2027.

\bibliography{aaai21}

\end{document}